\documentclass[sigconf]{acmart}
\usepackage{graphicx}

\usepackage{url} 
\usepackage{algorithm}
\usepackage{comment} 
\usepackage[noend]{algpseudocode} 
\usepackage{xspace} 
\floatstyle{ruled} 
\usepackage{amsmath} %for align
\usepackage{mathrsfs}
\usepackage{color}
\usepackage{graphicx}
\usepackage{booktabs} % for toprule midrule
\usepackage{makecell}
\usepackage{multirow}
\usepackage{subfigure}
 
\usepackage[T1]{fontenc} %% for true type 
\usepackage{aecompl}

\newcommand{\nthesection}{\arabic{section}}
\newcounter{theorem}
\newcounter{prop}
\renewcommand{\theprop}{\arabic{theorem}}
\newcounter{heu}
\renewcommand{\theheu}{\arabic{theorem}}
\newcounter{property}
\renewcommand{\theprop}{\arabic{theorem}}
\newcounter{lemma}
\renewcommand{\thelemma}{\arabic{theorem}}
\newcounter{cor}
\renewcommand{\thecor}{\arabic{theorem}}
\newenvironment{prop}{\begin{em}
        \refstepcounter{theorem}
        {\vspace{0ex}\noindent \bf Proposition \theprop:}}{
        \end{em}\eop\vspace{0ex}}%\hspace*{\fill}\vspace*{1ex}}

\newcounter{fact}
\renewcommand{\thefact}{\arabic{theorem}}
\newcounter{definition}[section]
\renewcommand{\thedefinition}{\nthesection.\arabic{definition}}

\newcounter{alg}[section]
\renewcommand{\thealg}{\nthesection.\arabic{alg}}

\newcounter{arule}
\renewcommand{\thearule}{\arabic{arule}}

\newcounter{claim}
\renewcommand{\theclaim}{\arabic{claim}}

\newcommand{\eop}{\hspace*{\fill}\mbox{$\Box$}}     % End of proof
\newcounter{example}
\renewcommand{\theexample}{\arabic{example}}

\newcommand{\eat}[1]{}

\newcounter{ccc}

\newcommand{\stitle}[1]{\vspace{0.0ex}\noindent{\bf #1}}
\newcommand{\etitle}[1]{\vspace{0.0ex}\noindent{\em \underline{#1}}}

\newcommand{\sstab}{\rule{0pt}{5pt}\\[-2.5ex]}
\newcommand{\bi}{\begin{itemize}}
\newcommand{\ei}{\end{itemize}}

\newcommand{\ie}{\emph{i.e.,}\xspace}
\newcommand{\eg}{\emph{e.g.,}\xspace}
\newcommand{\wrt}{\emph{w.r.t.}\xspace}

\newcommand{\kw}[1]{{\ensuremath {\mathsf{#1}}}\xspace}
\newcommand{\kwo}[1]{{\ensuremath {\mathsf{#1}}}}

\newcommand{\np}{{\kw{NP}}}
\newcommand{\aslsa}{\kwo{A}*\kw{LSa}}

\newcommand{\asbeam}{\kwo{A}*\kw{Beam}}
\newcommand{\mthged}{\kw{ged}}
\newcommand{\mthgedw}{\kw{\widetilde{ged}}}

\newcommand{\cbqp}{\kw{CBQP}}

\newcommand{\ged}{{\sc GED}\xspace}

\newcommand{\as}{{A*}\xspace}
\newcommand{\ourmtd}{{MATA*}\xspace}
\newcommand{\ourgnn}{{\sc SEGcn}\xspace}

\newcommand{\hun}{\kw{Hungarian}}
\newcommand{\vja}{\kw{VJ}}
\newcommand{\simgnn}{\kw{SimGNN}}

\newcommand{\gennas}{\kw{GENNA^*}}
\newcommand{\genn}{\kw{GENN}}

\newcommand{\greed}{\kw{GREED}}
\newcommand{\gmn}{\kw{GMN}}

\newcommand{\aids}{{\sc Aids}\xspace}
\newcommand{\imdb}{{\sc Imdb}\xspace}
\newcommand{\cancer}{{\sc Cancer}\xspace}

\AtBeginDocument{%
  \providecommand\BibTeX{{%
    \normalfont B\kern-0.5em{\scshape i\kern-0.25em b}\kern-0.8em\TeX}}}

\copyrightyear{2023}
\acmYear{2023}
\acmPrice{15.00}
\begin{document}

\title[\ourmtd: Learning to Match Nodes for  GED Computation ]{\ourmtd: Combining Learnable  Node Matching  with  \as Algorithm for Approximate  Graph Edit   Distance Computation}

% node mapping vs matching 


\author{Junfeng Liu}
\authornote{This work is mainly done  during an internship at Huawei.}
\affiliation{%
  \institution{Beihang University} 
  \city{Beijing}
  \country{China}
  } 
\email{liujunfeng@buaa.edu.cn}
% 0009-0006-8205-4564

\author{Min Zhou}
\affiliation{% 
  \institution{Huawei Noah's Ark Lab}
  \city{Shenzhen}
  \country{China}
  } 
\email{zhoumin27@huawei.com}
%0000-0002-4088-1266

\author{Shuai Ma}
\affiliation{%
  \institution{Beihang University} 
  \city{Beijing}
  \country{China}
  } 
\email{mashuai@buaa.edu.cn}
%0000-0002-4050-0443

\author{Lujia Pan}
\affiliation{%
  \institution{Huawei Noah's Ark Lab} 
  \city{Shenzhen}
  \country{China}
  } 
\email{panlujia@huawei.com}
%0000-0002-8988-4740
%Huawei Technologies Co., Ltd.

\renewcommand{\shortauthors}{Junfeng Liu, Min Zhou, Shuai Ma, \& Lujia Pan}

\begin{abstract}
Graph Edit Distance (\ged) is a general and domain-agnostic metric to measure graph similarity, widely used in graph search or retrieving tasks.
However, the exact \ged computation is known to be  \np-complete. For instance, the widely used \as algorithms explore the entire search space to find the optimal solution which inevitably suffers scalability issues.
Learning-based methods apply graph representation techniques to learn the \ged  by formulating a regression task, which can not recover the edit path and lead to inaccurate GED approximation  (\ie the predicted  \ged is smaller than the exact).
To this end, in this work, we present a data-driven hybrid approach \ourmtd for approximate \ged computation based on Graph Neural Networks (GNNs) and A* algorithms, which models from the perspective of learning to match nodes instead of directly regressing \ged.
Specifically, aware of the  structure-dominant operations (\ie node and edge insertion/deletion)   property in \ged computation, a structure-enhanced GNN is firstly designed to  jointly learn local and high-order structural information  for node embeddings for node matchings. 
Second, top-$k$ candidate nodes are produced  via a differentiable top-$k$ operation to enable the training for node matchings,  which is adhering to another property of \ged, \ie multiple optimal node matchings.
Third, benefiting from the candidate nodes, \ourmtd only performs on the promising search directions, reaching the solution efficiently.
Finally, extensive experiments  show the superiority of \ourmtd as it significantly outperforms the combinatorial search-based, learning-based and hybrid methods and scales well to large-size graphs.
\end{abstract}

%%%%%%%%%%%%%%%%%%%%%%%%%%%%%%%%%%%%%%%%%%%%%%%%%%%%%%%%%%%%%%%%%%%%%%%%%%%





\maketitle

\section{Introduction}
\label{sec-intro}

\begin{figure*}[tb!]
\centering
\includegraphics[width=1.4\columnwidth]{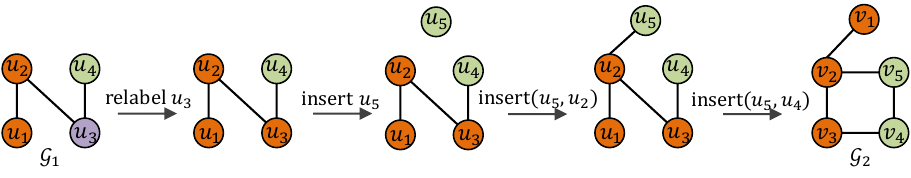} 
%\vspace{-2ex}
\caption{ 
An edit path from source graph $\mathcal{G}_1$ to target graph $\mathcal{G}_2$. Different colors represent the nodes with different labels.
Assume the edit costs are uniform, and \mthged$(\mathcal{G}_1, \mathcal{G}_2)=4$. 
That is at least four edit operations are required to transform $\mathcal{G}_1$ to $\mathcal{G}_2$, where the node mapping corresponding to the edit path is $\{u_1, u_2, u_3, u_4\}$  to $\{v_1, v_2, v_3, v_4\}$.
(1) Essentially, there are \emph{two optimal node matchings} for \mthged$(\mathcal{G}_1, \mathcal{G}_2)=4$, and another node mapping is $\{u_1, u_2, u_3, u_4\}$ to $\{v_2, v_3, v_4,v_5\}$. (2) Among the edit operations, there includes one attribute operation (\ie relabel $u_3$) and \emph{three structure operations}.} 
\label{fig:exam1}
%\vspace{-1ex}
\end{figure*}

Graph-structured data are ubiquitous, ranging from chemical compounds \cite{Carlos19Ligand}, social networks \cite{Fey20Deep}, computer vision \cite{Yan20Learning} to programming languages \cite{gmn}. 
A recurrent and pivotal task when working with these graph-structured applications is assessing how similar or different two given graphs are, among which graph edit distance (\ged) is a widely used metric due to its flexible and domain-agnostic merits \cite{gmn, lsa, bma, simgnn}.
In general, \ged computation refers to finding the minimum cost of edit operations (\emph{node insertion/deletion, edge insertion/deletion, and node/edge relabeling}) to transform the source graph to a target one \cite{David20Comparing} (see Fig.~\ref{fig:exam1} for an example).

The exact \ged computation guarantees optimality which is however \np-complete \cite{David20Comparing}.
It typically treats all possible edit operations as a pathfinding problem where A* algorithm (a best-first search method) is widely used to expand the search \cite{Kim19Inves, ChenHHV19, lsa, bma}.
These solutions mainly focus on pruning unpromising search spaces using A* algorithm or filtering dissimilar graph pairs to speed up \ged computation.
However, they all run in factorial time in the worst case due to the exhaustiveness of their search spaces, such that they cannot reliably compute the \ged  of graphs with more than $16$ nodes in a reasonable time~\cite{exact16nodes}.

Some recent works for the approximate \ged computation have been proposed with the help of the graph representation techniques, which can be divided into two main categories: Learning-based models \cite{gmn,simgnn,tagsim,graphsim,Peng2021Graph,h2mm, greed, zhuo2022efficient} and hybrid approaches \cite{genn,noah}.
For learning-based models, they directly formulate the approximate \ged computation as a regression task and supervisedly learn the \ged as a graph similarity metric in an end-to-end manner. Although such learning-based methods alleviate the computational burden of \ged, they could encounter the \emph{inaccurate GED approximation} issue (\ie the predicted \ged is smaller than the exact result) and also fail to recover an actual edit path, which is indispensable in specific tasks \eg network alignment \cite{koutra2013big}, graph matching \cite{cho2013learning, genn}. 
For hybrid methods, deep learning and combinatorial-search techniques are combined to optimize the \ged computation. 
Recently \cite{genn} and \cite{noah} separately propose two hybrid approaches, both of which apply Graph Neural Networks (GNNs) to guide the search directions of \as algorithms. 
However, the solved edit distance is typically provided with a large gap, due to their accumulation of the \emph{inaccurate GED approximation} in the cost function estimation (\ie the cost of unmatched subgraphs) of \as algorithms. 
Besides, GNNs with the attention mechanism are employed to estimate the cost function, which take $\mathcal{O}(n^2d + d^2n)$ time for extending each search, and encounter scalability issues~\cite{genn, noah}.

It is known that \ged computation equals finding the optimal node matching between the source and the target graphs. Once the node matchings are given,  \ged can  be easily calculated by  scanning the two graphs once \cite{lsa}, which reveals the intrinsic connection between \ged computation and node matching.
Besides, existing  learning-based and hybrid  approaches only formulate \ged  as a regression task for graph or subgraph pairs, which fails to  explicitly consider the node matching  in their models.
Be aware of the intrinsic connection between \ged computation and node matching, in this work, we attempt to learn the node matching corresponding to \ged using GNNs.
However, it is not trivial as the following two combinatorial properties essentially exist in \ged computation.
(1) \emph{Multiple optimal node matchings} (\ie different matchings to produce \ged) makes it difficult to learn the node matching  by directly modeling in end-to-end learning.
(2) \emph{Structure-dominant operations}  (\ie most edit operations are involved in structure) create challenges for incorporating structural information into learning models. Also, see Fig.~\ref{fig:exam1} for an example.

To this end, in this work, we present a data-driven  hybrid approach  {\textbf{MATA*}}  based on Graph Neural Networks and A* algorithms, which leverages the learned candidate \textbf{mat}ching nodes to prune unpromising search directions of \textbf{A*} algorithm (\ie~\aslsa~\cite{lsa}) for approximate \ged computation.

 \stitle{Contributions.} { The main contributions are as follows. }

\sstab  (1) We present a hybrid approach based on GNNs and A* algorithms rather than via an end-to-end manner, which models \ged computation from the perspective of node matching and combines the intrinsic connection between \ged computation and node matching.

\sstab  (2)  A structure-enhanced Graph Neural Network  (\ie~\ourgnn) is introduced to jointly learn local and high-order structural information  for node embeddings \wrt node matchings from a fine granularity, which  captures the combinatorial property of structure-dominant operations in \ged computation.

\sstab  (3) Further, top-$k$ candidate nodes are produced to be aware of the multiple optimal node matchings combinatorial property via a differentiable top-$k$ operation, which is built upon two  complementary learning tasks to jointly generate the candidate nodes, \ie learning node matching and learning \ged.

\sstab  (4) We conduct extensive experiments on real-life datasets \aids, \imdb, and \cancer to demonstrate the superiority and scalability of \ourmtd from three types of methods: combinatorial search-based, learning-based and hybrid approaches.
Indeed,  \ourmtd improves the accuracy by  (39.0\%, 21.6\%, 11.7\%) and reduces the average discrepancy  by (6.5\%, 9.1\%, 24.5\%) at least on three real-life datasets (\aids, \imdb, \cancer), respectively.

\section{Related Works}
\label{sec-intro}
Computing the graph edit distance between graphs is a classical and fundamental combinatorial optimization problem over graphs where a vast body of literature exists in various domains \cite{Riesen07Speeding,lsa, bma, Kim19Inves,asbeam, hunRiesenB09,vjFankhauserRB11, genn, noah, graphsim, tagsim, h2mm, gmn, greed, zhuo2022efficient}.
We next present a detailed overview of existing literature from three categories: (1) combinatorial search-based, (2) learning-based and (3) hybrid graph edit distance computation. 

\stitle{Combinatorial search-based.}
Combinatorial search-based algorithms either directly explore the search space corresponding to \ged, or relax it to other combinatorial problems with polynomial time complexity.
(1) The solution space of exact \ged is typically treated as a pathfinding problem where best-search (\as \cite{Riesen07Speeding, simpleas})  and depth-first search \cite{AbuAishehRRM15,BlumenthalG17} are utilized to  expand the search path \cite{noah}. 
Different exact algorithms mainly focus on how to better estimate the cost of unmatched subgraphs with the theoretical guarantee to prune the search space, such as using label sets \cite{Riesen07Speeding, RiesenEB13}, and subgraph structures \cite{lsa, bma, Kim19Inves}.
(2) The approximate algorithms are proposed to find the sub-optimal solutions. \cite{asbeam} explores the most possible directions with the limited beam size of A* algorithms. \cite{hunRiesenB09} and \cite{vjFankhauserRB11} only consider the local structure and relax it to bipartite matching problems, which are computed in cubic time. 
 
\stitle{Learning-based \ged computation.}
With the progress of graph representation techniques of Graph Neural Networks \cite{gcn, graphormer, Prakash2022Graph}, some works directly  model it as a regression problem and learn the approximate \ged via an end-to-end manner by treating \ged as a similarity score between graphs.
Different learning-based algorithms mainly focus on designing different GNN models for the graph edit distance computation task. 
\cite{simgnn} first presents a model using  GCN \cite{gcn} and attention mechanism to approximately learn \ged in an end-to-end fashion. 
Based on \cite{simgnn}, \cite{graphsim} further introduces a multi-scale node comparison technique to extract the fine-grained information from the node-to-node similarity matrix. 
Besides, \cite{gmn} incorporates both the node and graph level information by the cross-graph module to trade-off the accuracy and computation.
\cite{tagsim} splits the graph edit distance into different types of edit operations and applies graph aggregation layers to learn each type individually.
More recently, \cite{Peng2021Graph} designs a \ged-specific regularizer to impose the matching constraints involved in \ged, where the graph pairs are represented by the association graphs. 
\cite{greed} designs a novel siamese graph neural network, which through a carefully crafted inductive bias, learns the graph and subgraph edit distances via a property-preserving manner.

\stitle{Hybrid \ged computation.}
Recently, there has been a surge of interest in marrying learning-based approaches with combinatorial-search techniques. 
This interdisciplinary blend has given birth to several hybrid methodologies, particularly those that integrate Graph Neural Networks (GNNs) with the A* search algorithm, as seen in references~\cite{genn, noah}. 
Both methods leverage machine learning techniques to enhance the performance of  \as algorithms for \ged computation, by 
predicting the cost of unmatched subgraphs to optimize their search directions. 
\cite{noah} proposes graph path networks incorporating pre-training edit path information and cross-graph information for training the model and \cite{genn} integrates a  dynamic graph embedding network \cite{simgnn} for estimating the cost  associated with unmatched subgraphs.

\section{Preliminaries}
\label{sec-pre}

We focus the discussions on the labeled and undirected simple graphs, that is a graph denoted by  $\mathcal{G} = \{\mathcal{V, E,} {\Phi} \}$, where $\mathcal{V}$ is the set of nodes, $\mathcal{E}$ is the set of undirected edges with $\mathcal{E \subseteq V \times V}$ and ${\Phi}$ is a label function that assigns labels to each node or edge.

\stitle{Graph Edit Distance (\ged).}
The graph edit distance between graphs $\mathcal{G}_1$ and  $\mathcal{G}_2$ is defined as the minimum cost of edit operations (\ie \emph{node insertion/deletion, edge insertion/deletion, and node/edge relabeling}) to transform  $\mathcal{G}_1$ to $\mathcal{G}_2$, denoted by $ \mthged(\mathcal{G}_1,\mathcal{G}_2)$~\cite{tagsim, noah, bma}. 
One specific constraint to note is the aspect of \emph{node deletion} operation,  it's restricted only to the nodes that are isolated, ensuring structural integrity and meaningful transformations between the graphs.
Due to the NP-completeness of graph edit distance, the approximate edit distance is often used, denoted by $ \mthgedw(\mathcal{G}_1,\mathcal{G}_2)$, offers a balance between computational scalability  and accuracy. 
In this work, we focus on the line of uniform edit cost, \ie all of the edit operations share the same cost \cite{tagsim,simgnn,graphsim,lsa,greed,Peng2021Graph,bma,noah}, yet the techniques presented in the following sections can also be  extended to handle the non-uniform edit cost.

\stitle{GED computation from node matchings.}
We next illustrate how to compute \ged from the view of node matchings.  Here, the node matching refers to an injective function from the nodes $\mathcal{V}_1$ to  the nodes $\mathcal{V}_2$. 

\begin{prop}
\label{prop:mappings}
The \mthged  between the graph pair  $\mathcal{G}_1$  and $\mathcal{G}_2$ equals the minimum edit cost among all node matchings from the source graph $\mathcal{G}_1$ to the target graph $\mathcal{G}_2$ \cite{lsa}.
\end{prop}

By proposition \ref{prop:mappings}, the $ \mthged(\mathcal{G}_1,\mathcal{G}_2)$ can be determined by exhaustively generating all possible matchings from   $\mathcal{V}_1$  to   $\mathcal{V}_2$, \ie this essentially translates to identifying the node matching that incurs the least edit cost.  
Based on the commutativity of $\mthged( \cdot,\cdot)$ and the edit cost are uniform,  \emph{w.l.o.g.} for a graph pair $\mathcal{G}_1$ and $\mathcal{G}_2$, $\mathcal{G}_1$ always refers to the graph with fewer nodes in later sections.

Moreover, the optimal node matching between $\mathcal{G}_1$ and $\mathcal{G}_2$ can be formulated as the constrained binary quadratic programming (\cbqp) problem \cite{Peng2021Graph}:
\begin{equation}
\begin{aligned}\label{equ:cbqb}
\min _X \text { dist }= & \sum_{\substack{u_i \in \mathcal{G}_1 \\
v_k \in  \mathcal{G}_2}} c_{i, k} X_{i, k}+\sum_{\substack{u_i, u_j \in \mathcal{G}_1 \\
v_k, v_l \in \mathcal{G}_2}} c_{i, k, j, l} X_{i, k} X_{j, l} \\
\emph{ s.t. } & \sum_{v_k \in \mathcal{G}_2} X_{i, k}=1, \forall u_i \in \mathcal{G}_1 \\
& \sum_{u_i \in  \mathcal{G}_1} X_{i, k}=1, \forall v_k \in \mathcal{G}_2 \\
& X_{i, k} \in\{0,1\}, \forall u_i \in \mathcal{G}_1, v_k \in \mathcal{G}_2
\end{aligned}
\end{equation}
 
where $X \in |\mathcal{V}_1| \times  |\mathcal{V}_2|$ is a binary matrix  representing  the node matching between
$\mathcal{G}_1$ and $\mathcal{G}_2$. 
The value $X_{i,k}$ is 1 if node $u_i$ in $\mathcal{G}_1$ matches with node $v_k$ in $\mathcal{G}_2$.
The edit cost $ c_{i, k}$ denotes the cost of  matching $u_i$ in $\mathcal{G}_1$ and $v_k$ in $\mathcal{G}_2$. $ c_{i, k} = 1$ if  $u_i$ and $v_k$ have different labels and $0$ otherwise. Similarly, $c_{i, k, j, l}$ is the edit cost of matching the edge $(u_i,u_j)$ in  $\mathcal{G}_1$ and the edge $(v_k,v_l)$ in  $\mathcal{G}_2$. $c_{i, k, j, l}=1$ if  $(u_i,u_j)$ and $(v_k,v_l)$ have different labels and  $0$ otherwise.

\begin{table}[tb!]
\centering
\caption{Statistics of types of edit operations. We  randomly sample $1,000$ graphs for each dataset and compute their edit operations, where NI, EI  and ED stand for node insertions, edge insertions  and edge deletions, respectively. }
%\vspace{-2ex}
\resizebox{0.98\columnwidth}{!}
{
\begin{tabular}{l|ccc|c}
\toprule
\multirow{2}*{Datasets} & \multicolumn{3}{c|}{Structure Operations}         & Attribute Operations  \\ \cmidrule{2-5}
                  & \multicolumn{1}{c|}{NI} & \multicolumn{1}{c|}{EI} &  ED & Relabeling \\  \midrule
\aids            & \multicolumn{1}{c|}{$18.3$\%} & \multicolumn{1}{c|}{$34.8$\%}       & $8.7$\%        & $38.0$\%  \\
\imdb             & \multicolumn{1}{c|}{$12.1$\%} & \multicolumn{1}{c|}{$61.8$\%}       & $5.2$\%        & $0.0$\%   \\
\cancer         & \multicolumn{1}{c|}{$4.6$\%}  & \multicolumn{1}{c|}{$40.8$\%}       & $35.5$\%       & $18.6$\%  \\
\bottomrule
\end{tabular}
}
%\vspace{-1ex}
\label{tab:statis_operations}
%\vspace{-2ex}
\end{table}

\section{The Proposed Model: \ourmtd}
\label{sec-method}
%Different from formulating \ged computation as a regression task  via end-to-end learning, we design the hybrid approach \ourmtd, which models from the perspective of node matching and further incorporates the two combinatorial properties  (\ie {structure-dominant operations} and {multiple optimal node matchings}).   It employs  a structure-enhanced GNN (\ie~\ourgnn) to prune the unpromising search directions of an up-to-date combinatorial search-based algorithm \aslsa for approximate \ged computation.
 
Distinct from previous works that formulate \ged computation as a regression task, we suggest tracing the problem back to the node matching so that the  combinatorial properties  (\ie~{structure-dominant operations} and {multiple optimal node matchings}) in \ged computation could be leveraged.

\begin{figure*}[tb!]
\centering
 \includegraphics[width=0.86\textwidth]{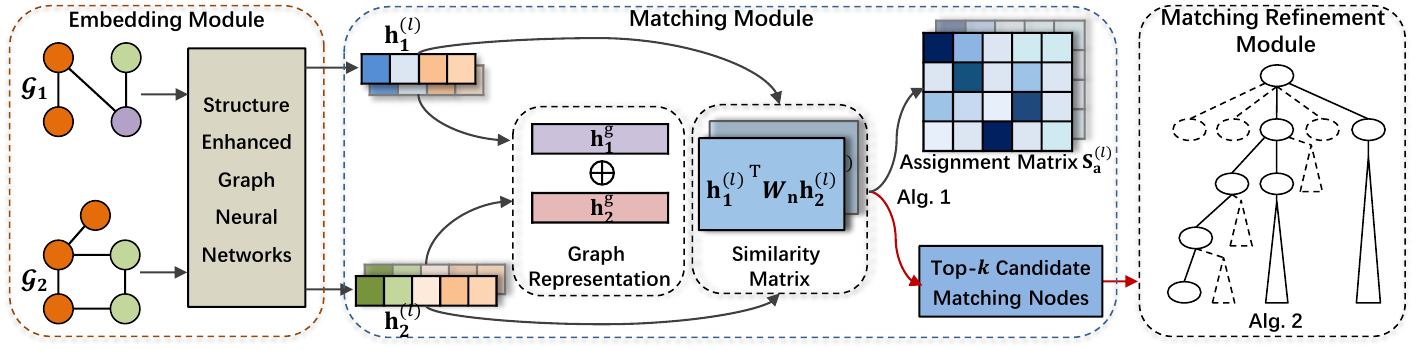}
%\vspace{-2ex}
\caption{The framework of \ourmtd. The black arrows stand for the data flow in the training and testing phases and the red arrows only denote that in the testing.   (1) Embedding module takes graph pairs as input and extracts the local and high-order  structural information via \ourgnn.
(2) Matching module utilized node embeddings to build two learning tasks, \ie  learning node matching with the help of the similarity matrix and learning GED using graph representation.
Further, top-$k$ candidates and the assignment matrix  are generated by Alg.~\ref{alg:topk}.   
(3) Benefiting from the candidate nodes, \ourmtd only performs on the promising search directions to refine these matchings using \aslsa. 
}
%\vspace{-2ex}
\label{fig:framework}
\end{figure*}

\subsection{Analysis of Learning to Match Nodes.}
   
In order to learn node matchings of \ged from \cbqp formulation, it requires   $X_{i, k} \in\{0,1\}$ to relax to be continuous in $[0, 1]$, where the constraints in  Eq.~(\ref{equ:cbqb}) could be modeled as the quadratic infeasibility penalty \cite{kochenberger2014unconstrained}. 
This relaxation strategy endows the binary matrix   $X$  with an augmented perspective. 
Thus, the  binary matrix $X$ can be viewed as the confidence of node $u_i$ in $\mathcal{G}_1$ matching with node $v_k$ in $\mathcal{G}_2$. 
In this way, we formulate the problem as the linear matching paradigm, by incorporating the graph structure information into node embedding, \ie solving the following transportation problem: 
\begin{equation}\label{equ:ot}
\min \sum_{i=1}^{|\mathcal{V}_1|} \sum_{j=1}^{|\mathcal{V}_2|} \mathbf{X}_{i j}\left\|\mathbf{h_1}_i-\mathbf{h_2}_j\right\|_2
\end{equation} 
where $\mathbf{h}_1 \in \mathbb{R}^{|\mathcal{V}_1|\times d}$ and $\mathbf{h}_2 \in \mathbb{R}^{|\mathcal{V}_2|\times d}$ are the node embeddings of $\mathcal{G}_1$ and $\mathcal{G}_2$, respectively. 
Intuitively, the kernel finds the optimal way to transform one set of node embeddings $\mathbf{h}_1$  to the other $\mathbf{h}_2$, by minimizing the Euclidean distance between corresponding node pairs \wrt graph pairs.

We further analyze  the combinatorial properties to design the approximate \ged computation framework, after modeling it from learning node matchings by solving Eq.~(\ref{equ:ot}). 
  
\stitle{Structure-dominant operations.}
Structure operations (\emph{node and edge insertion/deletion}) are dominant among all edit operations, which occupy at least  $62.0$\% as illustrated in Table~\ref{tab:statis_operations}. 
Indeed, the operations of \emph{node deletion} can be interpreted as \emph{node insertion}, as we arrange the source graph $\mathcal{G}_1$ and target $\mathcal{G}_2$ graph with $\mathcal{V}_1 \leq \mathcal{V}_2$~\cite{lsa}.
Further, by reducing the \cbqp formulation to the transportation problem in  Eq.~(\ref{equ:ot}), the graph structure information is assumed to be embedded into node embeddings.
These tell us we  need a  GNN to effectively learn powerful node embeddings enhanced by the graph structures (Section \ref{sec-method:emb}).

\stitle{Multiple optimal node matchings.}
Due to  the combination and permutation natures of node matchings, the two graphs typically have multiple optimal node matchings that yield the \ged.
Hence, directly learning the  node correspondence according to the matching confidence is extremely challenging, as it could lead to the inability to satisfy the injection constraint or a larger gap between $\mthged( \cdot,\cdot)$ and $\mthgedw( \cdot,\cdot)$. 
That is, it requires us to relax the constraint on the number of matched nodes and obtain  candidate nodes using a flexible parameter top-$k$. 
To conclude, we (1) need a differentiable top-$k$ operation to enable the  training  for node matchings (Section \ref{sec-method:mat}) and (2) refine the  matchings from  the top-$k$ candidates using \as algorithms (Section \ref{sec-method:find}).

Thus, the proposed \ourmtd employs  a structure-enhanced GNN (\ie~\ourgnn) to learn the differentiable top-$k$ candidate matching nodes which prunes the unpromising search directions of  \aslsa \cite{lsa} for approximate \ged computation. The overview of \ourmtd is illustrated in Fig.~\ref{fig:framework} with the details of each module elaborated in the following paragraphs.

\subsection{Embedding Module}
\label{sec-method:emb}
As analyzed, structural information is  critical to learn the fine-grained matching of node pairs for the \ged computation. 
Here, a structure-enhanced GNN \ourgnn is proposed that jointly learns the structural  information from the views of local and high-order. 
%\marked{Considering the computation efficiency as well as the properties of \ged task, node degree and position information are employed.}

\stitle{Degree encoding.}
When matching the nodes of two graphs \wrt~\ged, the nodes with similar degrees are more likely to be matched. 
%For example, the node $u_1$ of $\mathcal{G}_1$ matches $v_1$ of $\mathcal{G}_2$, as their degrees are both $0$ in Fig.~\ref{fig:exam1}.
Note that, the degree is not an accurate measure of structural similarity as the edge insertions and deletions are involved.
Hence, rather than directly encoding the degree with a one-hot vector, we assign each node with a learnable embedding {$d_i$} based on its degree, and the values of $d_i$  are randomly initialized.

\stitle{Position encoding.}
The nodes located with similar local positions are more likely to match in the \ged computation. Shortest-path-distances \cite{graphormer}, PageRank \cite{encodepagerank} and random walk  \cite{Prakash2022Graph,li20distance} are generally used to measure the relative position of nodes~\cite{yang2023kappahgcn}. For the sake of computation efficiency, we employ the probabilities of random walks after different steps as the relative position encoding $p_i \in \mathbb{R}^t $.
\begin{eqnarray}
\label{equ:rd}
p_i = [R_{ii}^{(1)},R_{ii}^{(2)}, \cdots,  R_{ii}^{(t)}],
\end{eqnarray}
where $R=AD^{-1}$ is the random walk operator, $t$ is the number of steps  of random walks, and $R_{ii}^{(t)}$ refers to  the landing probability of the node $i$ to itself after the $t$-$th$ step of random walk. Though the local positions are encoded via  random walk, nodes with slight structure distinction may hard to be  matched as the  Eq.~(\ref{equ:rd}) is deterministic yet the editing operations from  $\mathcal{G}_1$ to $\mathcal{G}_2$ do change the graph structures.

For more robust position encoding, perturbations are further injected to the original graphs to get the perturbed position encoding. Specifically, we  randomly insert and remove a small portion of edges (10\% utilized in experiments) to produce the perturbed graphs $\mathcal{G}_{in}$ and $\mathcal{G}_{re}$, respectively.
The random walk diffusion manners are further performed on $\mathcal{G}_{in}$ and $\mathcal{G}_{re}$ with the perturbed local positions  $p_i^{in}$ and  $p_i^{re}$, respectively. Combined with the Eq.~(\ref{equ:rd}), the positional encoding is given as follow:  
\begin{eqnarray}
\hat{p_i} = p_i + p_i^{in} + p_i^{re}.
\end{eqnarray}

\stitle{Local view.}
By concatenating (1) the node feature $x_i$ (\ie the attribute feature of its label), (2) degree encoding $d_i$, and (3) position encoding $ \hat{p_i}$, the node embeddings with local structure  $h_i^{(0)} \in  \mathbb{R}^d$ are built via a multilayer perceptron~(\texttt{MLP}):
\begin{eqnarray}
h_i^{(0)} = \texttt{MLP} (x_i \oplus d_i \oplus \hat{p_i} ), \ \ \forall i \in \mathcal{V}
\end{eqnarray}  
\stitle{High-order view.}
We adopt the GCN \cite{gcn} as the backbone of \ourgnn to learn the higher-order neighbor information.
The  node embeddings are aggregated from the embeddings of its adjacency nodes and itself.
The $l$-$th$ iteration of aggregation could be characterized as:
\begin{eqnarray}
h_i^{(l)} = \sigma \Big( \sum_{j\in \mathcal{N}_i}{\frac{1}{ c_{ij}} h_j^{(l-1)} w^{(l-1)}} \Big)
\end{eqnarray}
where $h_i^{(l)} \in  \mathbb{R}^d$ is the representation of node $i$  of $l$-$th$ GCN layer,  $\mathcal{N}_i$ is the set of neighbors of node $i$, and $w^{(l)}$ is the learned weights of $l$-$th$ layer.
In order to reduce the bias because of the different numbers of neighbors, the aggregated embeddings from adjacent nodes are also normalized by the total number of adjacent nodes $c_{ij}$. \ourgnn takes the obtained $h_i^0$  as the input embedding.

After encoding by \ourgnn, the node embeddings  of $\mathcal{G}_1$ from the local and high-order views are denoted as $\mathbf{h}_1^{(0)} \in \mathbb{R}^{|\mathcal{V}_1|\times d}$ and $\mathbf{h}_1^{(l)} \in \mathbb{R}^{|\mathcal{V}_1|\times d}$, respectively.
The node embeddings  of $\mathcal{G}_2$ are similarly obtained.

\subsection{Matching Module}
\label{sec-method:mat} 
The local and high-order structural affinities   between two graphs has been ingeniously  encoded into the node embedding space, by  \ourgnn.
As such, learning to match nodes is reduced to solve the Eq.~(\ref{equ:ot}).
We thus jointly learn the matchings from both the local view and high-order view to obtain the differentiable top-$k$ candidates  by iteratively minimizing the underlying transportation problem.
In addition to the task of learning node matching, a complementary  task learning \ged  is also put forward, which attempts to learn the distance between  graph representations that assists the node matching task.
 
\stitle{Learning node matchings.}
Intuitively, we learn node matchings from fine-grained correspondences to minimize the transportation problem, which aims that the resultant node matchings are not just approximations but are reflective of the genuine structural alignments between the two graphs.

\etitle{Similarity matrix.} % \in \mathbb{R}^{|\mathcal{V}_1|\times d}$\in \mathbb{R}^{|\mathcal{V}_2|\times d}$
To solve the Eq.~(\ref{equ:ot}), we  model it in a flexible way, and similarity matrices from the local view $\mathbf{S}^{(0)} \in |\mathcal{V}_1| \times  |\mathcal{V}_2|$ and the high-order view $\mathbf{S}^{(l)} \in |\mathcal{V}_1| \times  |\mathcal{V}_2|$ are: 
\begin{equation}\label{eqn:sim}
\begin{aligned} 
\mathbf{S}^{(0)} = \sigma ({{\mathbf{h}_1^{(0)}}^\top \mathbf{W_n}  \mathbf{h}_2^{(0)}}) \\
\mathbf{S}^{(l)} = \sigma ({{\mathbf{h}_1^{(l)}}^\top \mathbf{W_n}  \mathbf{h}_2^{(l)}}) 
\end{aligned}
\end{equation}  
where $\mathbf{W_n} \in \mathbb{R}^{d\times d}$ is a learnable weights matrix and shares the parameters between $\mathbf{S}^{(0)}$ and $\mathbf{S}^{(l)}$. 
All elements of similarity matrix $\mathbf{S}$ are positive after applying the sigmoid function, and $\mathbf{S}^{(0)}_{i,j}$ measures the similarity between $\mathcal{V}_{1i}$ and $\mathcal{V}_{2j}$ from the local view. And $\mathbf{S}^{(l)}_{i,j}$ measures the similarity from the high-order view.
Besides, $\mathbf{S}^{(0)}$ also models the cost by transforming the embedding $\mathbf{h}_1^{(0)}$ to $\mathbf{h}_2^{(0)}$.
Different from  padding  the similarity matrix \cite{graphsim}, it is  enough to represent  all possible matchings with $|\mathcal{V}_1| \times  |\mathcal{V}_2|$ from the Eq.~\ref{equ:cbqb}.

%%%%%%%%%%%%%%%%%%%%%%%%%%%%%%%%%%%%%%%%%%%%%%%%%%%%%%%%%%%%%%%%%%%%%%%%%
\begin{algorithm}[tb!] 
\caption{Differentiable top-$k$ matching nodes} \label{alg:topk} 
\hspace{-7ex} \textbf{Input:}  Similarity mat $\mathbf{S}^{(0)}$, $\mathbf{S}^{(l)}$, $k$, regularization $\epsilon$  \\
\hspace{-1ex} \textbf{Output:}  Assignment mat $\mathbf{S_a}^{(0)}$, $\mathbf{S_a}^{(l)}$,  candidates  $M^{|\mathcal{V}_1| \times k}$
\begin{algorithmic}[1]
%\State Compute the matching order \mo of $\mathcal{V}_1$;
\State  Build $\mathbf{D}$, $\mathbf{c}$, $\mathbf{r}$ from $\mathbf{S}^{(0)}$ by Eq.~(\ref{equ:flatten}); $\mathbf{\Gamma}= -\mathbf{D}/\epsilon$;
\While{$\mathbf{\Gamma}$ is not converged}  \Comment{Sinkhorn normalization}
\State $\mathbf{\Gamma}=\operatorname{diag}((\mathbf{\Gamma} \mathbf{1} \odot \mathbf{r}))^{-1} \mathbf{\Gamma}$;
\State $\mathbf{\Gamma}=\operatorname{diag}\left(\left(\mathbf{\Gamma}^{\top} \mathbf{1} \odot \mathbf{c}\right)\right)^{-1} \mathbf{\Gamma}$;
\EndWhile
\State Rebuild assignment mat $\mathbf{S_a}^{(0)}$ from $\mathbf{\Gamma}$;
\State Repeat lines 1--5 for $\mathbf{S}^{(l)}$ to obtain   $\mathbf{S_a}^{(l)}$;
\If{ training} { \textbf{return} $\mathbf{S_a}^{(0)}$, $\mathbf{S_a}^{(l)}$ ;}
\Else \ { \textbf{return} $M^{|\mathcal{V}_1| \times k}$ by greedily searching top-$k$;}
\EndIf
\end{algorithmic}
\end{algorithm}

%%%%%%%%%%%%%%%%%%%%%%%%%%%%%%%%%%%%%%%%%%%%%%%%%%%%%%%%%%%%%%%%%%%%%%%%%

\etitle{Top-$k$ candidate matching nodes.}
Inspired by \cite{xie2020differentiable, wang2023deep}, choosing the top-$k$ matches from the  similarity matrices $\mathbf{S}^{(0)}_{i,j}$ and $\mathbf{S}^{(l)}_{i,j}$ is typically formulated as an optimal transport problem, which  selects the $k$ most confident matches for each node based on the  matching confidences, shown in Alg.~\ref{alg:topk}.

Specifically, we first flatten the similarity matrix $\mathbf{S}^{(0)}$ with  local structure affinity into  $\mathbf{d}=[d_1, d_2, ..., d_{|\mathcal{V}_1||\mathcal{V}_2|}]$.
As such,  to differentiable find the top-$k$ matches, 
 the optimal transport problem can be viewed as redistributing $\mathbf{d}$ to one of  $d_{max}$ and $d_{min}$, where  the capacities of  $d_{max}$ and $d_{min}$ are $k$ and $|\mathcal{V}_1||\mathcal{V}_2|-k$, respectively.
That is the matches moved into $d_{max}$ are preserved during the redistributing and the others moved into  $d_{min}$ are discarded. 
Let  $\mathbf{c}$ and  $\mathbf{r}$ represent the marginal distributions, $\mathbf{D}$  represents the distance matrix, and $\mathbf{1}$ represents the vector of all ones (line 1). And we have:
\begin{equation}
\begin{aligned} \label{equ:flatten}
\mathbf{r} & =\mathbf{1}_{|\mathcal{V}_1||\mathcal{V}_2|}^{\top}, \quad \mathbf{c}=\left[|\mathcal{V}_1||\mathcal{V}_2|-k, k\right]^{\top} \\
\mathbf{D} & =\left[\begin{array}{llll}
d_1-d_{\min } & d_2-d_{\min } & \cdots & d_{|\mathcal{V}_1||\mathcal{V}_2|}-d_{\min } \\
d_{\max }-d_1 & d_{\max }-d_2 & \cdots & d_{\max }-d_{|\mathcal{V}_1||\mathcal{V}_2|}
\end{array}\right]
\end{aligned}
\end{equation}

Then an efficient  method Sinkhorn \cite{sinkhorn, Fey20Deep} for solving the optimal transport problem is typically adopted  to learn the probabilities of the top-$k$ matchings,  which is an approximate and differentiable version of \hun. It  iteratively performs row-normalization, \ie element-wise division by the sum of its row and column-normalization  until convergence, where $\odot$ means element-wise division, $\operatorname{diag}(\cdot)$ means building a diagonal matrix from a vector (lines 2--4).

After the differentiable top-$k$ operation, we  reshape $\mathbf{\Gamma}$ into the \emph{assignment matrix} from local view $\mathbf{S_a}^{(0)} \in |\mathcal{V}_1|\times|\mathcal{V}_2|$  , which   essentially measures the confidence  of $\mathcal{V}_{1i}$ and $\mathcal{V}_{2j}$ belonging to the optimal matching (line 5).
For the  similarity matrix $\mathbf{S}^{(l)}$ with  high-order structure affinity are also performed to obtain the \emph{assignment matrix} $\mathbf{S_a}^{(l)}$ from high-order view   (line 8).
Finally, Alg.~\ref{alg:topk}  returns  $\mathbf{S_a}^{(0)}$, $\mathbf{S_a}^{(l)}$ during the training, and top-$k$ candidate nodes $M^{|\mathcal{V}_1| \times k}$ during the testing (lines  6--7).

Note that,  during the testing, we further propose a greedy method to find top-$k$ candidate nodes  in $\mathcal{O}(kn^2)$ time, In brief, it iteratively finds a node with the largest matching probability as a candidate  from the \emph{unmatched nodes}, where the injection constraint of node matchings is also guaranteed.

\stitle{Learning \ged.}
We further propose an auxiliary task  tailored  to learn the   (approximate) graph edit distance that assists the node matching task by exploiting the graph-level similarity.

%%%%%%%%%%%%%%%%%%%%%%%%%%%%%%%%%%%%%%%%%%%%%%%%%%%%%%%%%%%%%%%%%%%%%%%%%
\begin{algorithm}[tb!] 
\caption{Mapping refinement based on \aslsa [1]} \label{alg:lsa} 
\hspace{-14ex} \textbf{Input:} Graphs $\mathcal{G}_1$, $\mathcal{G}_2$, candidates $M^{|\mathcal{V}_1| \times k}$ \\
\hspace{-18ex} \textbf{Output:}  The approximate $\mthgedw(\mathcal{G}_1,\mathcal{G}_2)$ 
\begin{algorithmic}[1]
%\State Compute the matching order \mo of $\mathcal{V}_1$;
\State Push ${( 0, \emptyset, null, 0)}$ into $Q$;  \hspace{4ex} \Comment{Initialize the priority queue $Q$ with the root of the search tree.}
\While{$Q \neq \emptyset$}
\State Pop $(i, f, pa, lb)$ with minimum $lb$ from $Q$;
\State Compute the lower bound $lb$ using \aslsa for each child $c \in M^{|\mathcal{V}_1| \times k}$   of $f$;
\ForAll {child $c \in M^{|\mathcal{V}_1| \times k}$  of $f$ } %with $c.lb < \mathcal{T}$
\If{$ i + 1 = |\mathcal{V}_1|$}  {$\mthgedw(\mathcal{G}_1,\mathcal{G}_2)$ = $c.lb$;  \ \textbf{break;} }
\Else \ {Push $(i+1, c, f, lb)$ into $Q$;}
\EndIf
\EndFor
\EndWhile \\
\textbf{return} $\mthgedw(\mathcal{G}_1,\mathcal{G}_2)$
\end{algorithmic} 
\end{algorithm}%\vspace{-2ex}
%%%%%%%%%%%%%%%%%%%%%%%%%%%%%%%%%%%%%%%%%%%%%%%%%%%%%%%%%%%%%%%%%%%%%%%%%

Intuitively, $\mathbf{h}_1^{(0)}$ and  $\mathbf{h}_1^{(l)}$  capture the node  features enhanced by  the local and high-order graph structural information of $\mathcal{G}_1$.
This intricate embedding process ensures that the nodes' features are not only captured in their raw features but are also contextualized within the broader structure of the graph.
Essentially, (approximate) \ged measures the similarity of graph pairs from the graph level, and hence, we  aggregate the node embeddings with both local and high-order views  of  $\mathcal{G}_1$ and  $\mathcal{G}_2$  passed through  $\texttt{MLPs}$ for the  learning \ged task. 
And we  have the following:
 \begin{equation} %\mathbf{h}^g
 \begin{aligned}
 \mathbf{h}_{1}^g =  \texttt{MLP} ( \mathbf{h}_1^{(0)}  \oplus \mathbf{h}_1^{(l)} )  \\
 \mathbf{h}_{2}^g =  \texttt{MLP} ( \mathbf{h}_2^{(0)}  \oplus \mathbf{h}_2^{(l)} )  \\
d_{\mathcal{G}_1,\mathcal{G}_2} = \texttt{MLP} ( \mathbf{h}_{1}^g \oplus \mathbf{h}_{2}^g )     
\end{aligned}
\end{equation}
That is,
$d_{\mathcal{G}_1,\mathcal{G}_2}$ is predicted using the $\texttt{MLP}$ operation which gradually reduces the concatenated graph representations $\mathbf{h}_{1}^g$ and $\mathbf{h}_{2}^g$ of the graph pair.
Actually, to counter this and provide a more interpretable and standardized measure, the \ged are typically normalized by $\exp \{-\mthged( \mathcal{G}_1, \mathcal{G}_2 ) \times 2 / (\mathcal{V}_1 + \mathcal{V}_2) \}$.

\stitle{Loss design.}
\ourmtd is trained in a supervised manner for graph pairs $\mathcal{G}_x$ and $\mathcal{G}_y$ using normalized ground-truth \ged ${d}^{t}_{x,y}$   and its corresponding node matching $\mathcal{M}^{t}_{x,y}$.
The loss function evaluates both the difference for predicted node matchings from the local/high-order view of assignment matrices $\mathbf{S_a}^{(0)}$/$\mathbf{S_a}^{(l)}$ and learning \ged from the predicted normalized \ged  $d_{x,y}$.
For the learning node matching task, we jointly minimize the negative log-likelihood of the node matchings on the assignment matrices $\mathbf{S_a}^{(0)}$ and $\mathbf{S_a}^{(l)}$:
\begin{eqnarray}
\mathcal{L}_n = - \frac{1}{|\mathcal{D}|}  \sum_{(x,y)\in \mathcal{D}} \sum_{(i,j)\in \mathcal{M}^{t}_{x,y}}{  \texttt{log} \mathbf{S_a}^{(0)}_{i,j} + \texttt{log} \mathbf{S_a}^{(l)}_{i,j}}
\end{eqnarray} % \mathbf{S_a^t}
Note that, different from the use of permutation cross-entropy loss \cite{perloss} or Hungarian loss \cite{hunloss} for the graph matching task, only the node pairs belonging to a node matching are penalized by $\mathcal{L}_n$, the other node pairs are not penalized.
The rationale behind this is that multiple optimal node matchings typically exist, and these unmatched node pairs may also belong to other node matchings corresponding to the \ged.

For the learning \ged task, we minimize the MSE loss:
\begin{eqnarray}
\mathcal{L}_g = \frac{1}{|\mathcal{D}|}  \sum_{(x,y)\in \mathcal{D}}{ ( d_{x,y} - d^{t}_{x,y} )^2}
\end{eqnarray} %d(\mathcal{G}_x,\mathcal{G}_y) %$\mathcal{G}_x,\mathcal{G}_y$.
where $\mathcal{D}$ is the set of training graph pairs. 

Our final loss function is a combination of the  negative log-likelihood loss and MSE loss:  $ \mathcal{L} = \mathcal{L}_g + \mathcal{L}_n$.

%Optimal Finding Module
\subsection{Mapping Refinement Module}
\label{sec-method:find}
\ourmtd finally integrates  \aslsa algorithm \cite{lsa} to refine the edit distance (\ie node matching)  among the learned top-$k$ candidate matching nodes $M^{|\mathcal{V}_1| \times k}$, as shown in Alg.~\ref{alg:lsa}. 

Specifically, \ourmtd   conducts a best-first search by treating \ged computation  as a pathfinding problem. 
Such a representation is convenient because it provides a systematic and heuristic way to explore possible node matchings. 
To facilitate this search process, a priority queue  is maintained to store the search states  during the process for guiding the search direction.
The priority queue $Q$ contains the level $i$, current partial matching $f$, the parent matching $pa$, and the lower bound $lb$.
\aslsa  initializes the priority queue $Q$ by the root of the search tree (line 1).
It then iteratively pops $(i, f, pa, lb)$ from $Q$ with the minimum lower bound, and subsequently  extends the current matching $f$ by computing the lower bound of each child  belonging to the candidates $M^{|\mathcal{V}_1| \times k}$ (lines 2--7).
If the full node matching is formed, then $\mthgedw(\cdot,\cdot)$ equals  its lower bound and is returned (lines 6, 8).

Note that, different from the hybrid approach~\cite{noah,genn}, during the mapping refinement of \ourmtd, the search space  is  pruned by the theoretical bounded estimation of unmatched subgraphs of \aslsa.

\section{Experiments}
\label{sec-exp}

\subsection{Experimental Settings}

 %%%%%%%%%%%%%%%%%%%  dataset statistic
\begin{table}[tb!]
\centering
\setlength{\tabcolsep}{4pt}
\renewcommand{\arraystretch}{1.15}  
\caption{Statistics of datasets. The graph pairs are partitioned 60\%, 20\%, 20\% as training, validation, test sets, respectively.}\vspace{-2ex}
\resizebox{0.98\columnwidth}{!}
{
\begin{tabular}{lcccccc}
\toprule
       & |Graphs| & |Pairs|  & avg($\mathcal{\frac{|E|}{|V|}}$)  & min($\mathcal{|V|}$) & max($\mathcal{|V|}$) & avg($\mathcal{|V|}$) \\ \midrule
\aids   & $700$    & $490$K      & $0.98$  & $2$      & $10$     & $8.90$     \\
\imdb   & $1500$   & $2.25$M     & $4.05$  & $7$      & $89$     & $13.00$    \\
\cancer & $800$    & $100$K      & $1.08$    & $21$     & $90$     & $30.79$  \\
\bottomrule
\end{tabular}
}
\label{tab:statis_dataset}
%\vspace{-2ex}
\end{table}
%%%%%%%%%%%%%%%%%%%

\stitle{Datasets.}
In this work, three benchmark datasets \ie~\aids~\cite{simgnn}, \imdb~\cite{imdb}, and \cancer~\footnote{https://cactus.nci.nih.gov/download/nci/CAN2DA99.sdz \label{cancer}} are employed.
(1) \aids is a set of antivirus screen chemical compounds labeled with $29$ types.  Following \cite{simgnn,genn},  $700$ graphs with no more than ten nodes are sampled as the \aids dataset.
(2) \imdb consists of $1,500$ ego-networks of movie actors or actresses and each of which is an non-attributed graph.
(3) \cancer consists of $32,577$ graphs of molecules discovered in carcinogenic tumors. To test the scalability and efficiency of our \ourmtd, we sample $800$ graphs with nodes from $21$ to $90$ as \cancer dataset, where the nodes are labeled with $37$ types of atoms.
Statistics of the three real-life datasets are shown in Table \ref{tab:statis_dataset}.

%%%%%%%%%%%%%%%%%%% Solution quality evaluations
\begin{table*}[tb!]
\centering
\caption{Effectiveness evaluations. The metrics are calculated on the normalized edit distance. $\uparrow$ indicates  the high the better and   $\downarrow$ otherwise.  Top-$k$ are set to $4$, $6$ and $8$ for \aids, \imdb and \cancer, respectively. The unit of metrics ACC and MSE are \% and $10^{-2}$, respectively, and  -- refers to memory overflow on 32GB machines or runs in more than 10 minutes for one graph pair. }
%\vspace{-2ex}
\setlength{\tabcolsep}{11pt}
\renewcommand{\arraystretch}{1.15}
\resizebox{0.95\textwidth}{!}
{
\begin{tabular}{llcccccccc}
\toprule
 \multicolumn{1}{l}{Datasets}                   & \multicolumn{1}{l}{ Methods}                       &  \makecell[c]{Edit \\ Path}  & ACC $\uparrow$ & MAE $\downarrow$ & MSE $\downarrow$ & p@10 $\uparrow$  & p@20 $\uparrow$ & $\rho$ $\uparrow$ & $\tau$ $\uparrow$ \\ \midrule
\multirow{8}*{\aids} % MSE $(10^{-2})$ ACC(\%)
&\asbeam \cite{asbeam}        & $\checkmark$  & $16.68$     & $0.092$     & $1.37$    & $0.460$     & $0.470$     & $0.720$     & $0.546$     \\
&\hun \cite{hunRiesenB09}     & $\checkmark$  & $4.19$     & $0.194$     & $4.77$     & $0.293$     & $0.328$     & $0.541$     & $0.386$     \\
&\vja \cite{vjFankhauserRB11} & $\checkmark$  & $0.95$     & $0.216$     & $5.64$     & $0.215$     & $0.273$     & $0.543$     & $0.387$     \\ \cmidrule{2-10}
&\simgnn  \cite{simgnn}       & $\times$      & $0.01$      & $0.036$     & $0.22$     & $0.470$     & $0.540$     & $0.886$     & $0.725$   \\
&\gmn \cite{gmn}              & $\times$      & $0.02$       & $0.034$       & $0.19$     & $0.401$       & $0.489$       & $0.750$       & $0.673$      \\
&\greed \cite{greed}              & $\times$  & $0.00$     & $0.031$     & \textbf{0.17}    & $0.461$     & $0.533$     & $0.894$     & $0.732$  \\
&\genn \cite{genn}            & $\times$      & $0.02$      & $0.031$ & \textbf{0.17}     & $0.441$     & $0.525$     & \textbf{0.898}     & \textbf{0.738}     \\   \cmidrule{2-10}
&\gennas \cite{genn}          & $\checkmark$  & $20.05$     & $0.034$     & $0.46$     & $0.407$     & \textbf{0.556}     & $0.515$     & $0.378$    \\  %\midrule
%&\ourmtd (Ours)  & $\checkmark$  & \textbf{65.06} & \textbf{0.027}     & $0.32$     &\textbf{0.536}     & \textbf{0.565}     & $0.855$     & $0.721$     \\
&\ourmtd (Ours)  & $\checkmark$  & \textbf{59.12} & \textbf{0.029}     & $0.37$     &\textbf{0.486}     & {0.526}     & $0.844$     & $0.698$     \\
\midrule

\multirow{8}*{\imdb}
&\asbeam \cite{asbeam}        & $\checkmark$  & $23.18$     & $0.111$     & $5.22$     & $0.464$     & $0.527$     & $0.489$     & $0.381$    \\
&\hun \cite{hunRiesenB09}     & $\checkmark$  & $22.53$     & $0.115$     & $5.38$     & $0.438$     & $0.498$     & $0.465$     & $0.359$    \\
&\vja \cite{vjFankhauserRB11} & $\checkmark$  & $22.24$     & $0.115$     & $5.38$     & $0.436$     & $0.495$     & $0.465$     & $0.359$     \\ \cmidrule{2-10}
&\simgnn   \cite{simgnn}      & $\times$      & $0.11$     & $0.114$     & \textbf{5.01}     & $0.474$     & $0.531$     & $0.500$     & $0.388$    \\
&\gmn \cite{gmn}              & $\times$      & $0.29$     & $0.128$     & \textbf{5.01}     & $0.479$     & $0.542$     & $0.513$     & $0.392$    \\
&\greed \cite{greed}        &$\times$  & $0.93$     & $0.110$     & $5.04$    & $0.477$     & $0.541$     & $0.499$     & $0.389$     \\
&\genn \cite{genn}            & $\times$      & $0.22$     & $0.108$     & $5.04$     & $0.476$     & $0.533$     & $0.495$     & $0.384$     \\   \cmidrule{2-10}
&\gennas \cite{genn}          & $\checkmark$  &  --      &  --      & --     & --       & --       &  --      & --       \\  %\midrule
&\ourmtd (Ours)               & $\checkmark$  & \textbf{44.80}     & \textbf{0.098}     & {5.03}     & \textbf{0.509}     & \textbf{0.570}     & \textbf{0.542}     & \textbf{0.456}     \\
\midrule

\multirow{8}*{\cancer} %%%%%%% k=8
&\asbeam\cite{asbeam}         & $\checkmark$  & $44.23$    & $0.053$    & {1.14}    & $0.161$    & $0.266$    & $0.446$    & $0.352$     \\
&\hun \cite{hunRiesenB09}     & $\checkmark$  & $2.19$    & $0.162$    & $3.56$    & $0.123$    & $0.227$    & $0.139$    & $0.096$    \\
&\vja \cite{vjFankhauserRB11} & $\checkmark$  & $0.00$    & $0.184$    & $4.85$    & $0.095$    & $0.187$    & $0.188$    & $0.133$   \\ \cmidrule{2-10}
&\simgnn \cite{simgnn}        & $\times$      & $0.01$    & $0.068$    & $1.42$    & $0.273$    & $0.297$    & $0.277$    & $0.191$    \\
&\gmn \cite{gmn}              & $\times$      & $0.00$    & $0.071$    & $1.47$    & $0.280$    & $0.285$    & $0.254$    & $0.174$   \\
&\greed \cite{greed}  & $\times$  & $0.00$     & $0.077$     & $1.86$    & $0.131$     & $0.164$     & $0.170$     & $0.118$ \\
&\genn \cite{genn}            & $\times$      & $0.00$    & $0.069$    & $1.44$    & $0.285$    & $0.264$    & $0.300$    & $0.207$     \\   \cmidrule{2-10}
&\gennas \cite{genn}          & $\checkmark$  &  --      &  --      & --     & --       & --       &  --      & --    \\  %\midrule
&\ourmtd (Ours)               & $\checkmark$  & \textbf{55.89}    & \textbf{0.040}    & \textbf{1.13}    & \textbf{0.820}    & \textbf{0.825}    & \textbf{0.729}    & \textbf{0.625}    \\

\bottomrule
\end{tabular} 
}
\label{tab:quality}%\vspace{-2ex}
\end{table*}
%%%%%%%%%%%%%%%%%%%

\stitle{Baseline methods.}
Our baselines include three types of methods, combinatorial search-based algorithms, learning-based models and hybrid approaches.
(1) The representative methods in the first category include three well-known approximate algorithms \asbeam \cite{asbeam}, \hun \cite{hunRiesenB09} and \vja \cite{vjFankhauserRB11}.
(2) The second category includes three common-used and one state-of-the-art learning models, \ie~\simgnn~\cite{simgnn}, \gmn~\cite{gmn}, \genn~\cite{genn} and \greed~\cite{greed}.
(3) We chose an up-to-date model \gennas as the representative of hybrid approaches, and our  \ourmtd also belongs to this category.

%(1) Edit path (\ie whether a method can recover the edit path.)
%(2) Accuracy (ACC), (3) {Mean Absolute Error} (MAE), (4) {Mean Squared Error} (MSE),  (5) {Precision at $k$} (p@10), and (6) p@20 are the commonly used metrics.
%(7) {Spearman’s Rank Correlation Coefficient} ($\rho$) and (8) {Kendall’s Rank Correlation} ($\tau$), both of which measure how well the computed results match with the ground-truth ranking results.
%(9) {Time}, which records the average running time per graph pair. %Note that the ground-truth is normalized by Equation \ref{equ:nor-ged}. More details of the evaluation metrics are recorded in Appendix \ref{sec-app:exp}.

\stitle{Evaluation metrics.}
We adopt the following experimental metrics to evaluate the performance of the various approaches:
(1) {Edit path} means whether a method can recover the edit path  corresponding to the computed edit distance. 
(2) {Accuracy} (ACC), which measures the accuracy between the computed distance and the ground-truth solutions. 
(3) {Mean Absolute Error} (MAE), which indicates the average discrepancy between the computed distance and ground truth. 
(4) {Mean Squared Error} (MSE), which stands for the average squared difference between the computed distance and ground truth.
(5) {Precision at $10$} (p@10) and (6) {Precision at $20$} (p@20), both of which mean the precision of the top $10$ and $20$ most similar graphs within the ground truth top $10$ and $20$ similar results. 
(7) {Spearman’s Rank Correlation Coefficient} ($\rho$) and (8) {Kendall’s Rank Correlation} ($\tau$), both of which measure how well the computed results match with the ground-truth ranking results. 
(9) {Time}. It  records the running time to compute the distance for one graph pair. The  methods involving learning only report the testing time.

Due to the fact that exact \ged computation is \np-complete, the ground-truth of \aids is produced by exact algorithms and the ground-truth of \imdb and \cancer are generated by the smallest edit distances of \asbeam, \hun, and \vja, following \cite{simgnn}.
Note that, \ourmtd is able to achieve a smaller edit distance, and the ground-truth of \imdb and \cancer are further updated by the best results of the four approaches. Therefore, the metrics on \aids are calculated by the exact solutions and the metrics on \imdb and \cancer are calculated by the updated ground-truth. 
Note that the edit distance is normalized into a similarity score in the range of $(0,1]$ as explained in Section~\ref{sec-method:mat},  the same as~\cite{simgnn,genn}.
% \begin{equation}\label{equ:nor-ged}
% \exp \{-\mthged( \mathcal{G}_1, \mathcal{G}_2 ) \times 2 / (\mathcal{V}_1 + \mathcal{V}_2) \},
% \end{equation} 

\stitle{Parameter settings.}
We  conduct all experiments on machines with Intel Xeon Gold@2.40GHz CPU and NVIDIA Tesla V100 32GB GPU. 
The number of \ourgnn layers, \ie $l$ is set to $3$ and the random walk step $t$ is set to $16$ for the three datasets.
During the training, we set the batch size to 128 and use Adam optimizer  with 0.001 learning rate and $5\times 10^4$ weight decay for each dataset. The source codes and data are available at \url{https://github.com/jfkey/mata}.

\subsection{ Experimental Results}
%\marked{In this section, we evaluate the performance of approximate \ged computation and give the analyses of \ourmtd. More detailed experiments \wrt top-$k$ section are demonstrated in Appendeix \ref{sec-app:exp}. }
In this section, we evaluate the performance of \ourmtd from the effectiveness, scalability, efficiency, ablation study, top-$k$ comparisons, and the analysis of assignment matrices.
%More experiments about top-$k$ comparisons are demonstrated in Appendeix \ref{sec-app:exp}.

\stitle{Effectiveness evaluations.}
Table \ref{tab:quality} shows the effectiveness of eight approaches on three real-world datasets.
\ourmtd consistently achieves the best performance under almost each evaluation metric, which demonstrates the superiority of our hybrid method \ourmtd incorporating the two combinatorial properties of \ged computation.
We conduct the following findings from the  evaluations.

(1) From the ACC, \ourmtd achieves smaller edit distances at least (58.1\%, 32.1\%, 53.6\%) of graph pairs on (\aids, \imdb, \cancer) when comparing with combinatorial search-based and hybrid approaches. Hence, the ground-truth of \imdb and \cancer are further updated by these, which reduces MAE by at least (6.5\%, 9.1\%, 24.5\%).  
(2) Only learning-based models cannot recover the edit path, as they directly learn \ged as a similarity score and ignore the combinatorial nature.
(3) On \imdb, all methods perform worse than on \aids and \cancer.
\imdb is large with a wide range from $7$ to $89$ nodes. Besides, the graphs are much denser with $\mathcal{|E|}/\mathcal{|V|}$ = $4.05$ and the distances of pairs are also larger, which increases the difficulty for combinatorial search and learning methods.

(4) The improvement of \ourmtd on  ACC is such significant with at least (39.0\%, 21.6\%, 11.7\%), and the improvement of other metrics is relatively less significant. The rationales behind this lie in that (a) \ourmtd models from the perspective of node matching   and explicitly build the task of learning node matching, that is the learned top-$k$ candidate nodes could directly improve the accuracy due to the correspondence between \ged and the node matching.  (b) For fewer node pairs whose matchings have not been learned by \ourmtd, it prunes the search subtree rooted at the these node pairs, which leads to a larger edit distance reflected in other metrics.

%%%%%%%%%%%%%%%%%%%%% Running time
\begin{table}[tb!]
\centering
\caption{Efficiency evaluations. Average running time for solving one graph pair on test data (\emph{ms}). The training time for learning-based and hybrid approaches does not include.}
\setlength{\tabcolsep}{2pt}
%\vspace{-2ex}
\renewcommand{\arraystretch}{1.15}
\resizebox{0.98\columnwidth}{!}
{
\begin{tabular}{lccc|ccc|cc}
\toprule
       & \simgnn & \gmn  & \genn & \asbeam  & \hun   & \vja    & \gennas  & \ourmtd \\ \hline
\aids   & \textbf{0.3}    & $9.0$  & $0.4$  & $20.4$  & $6.7$   & $6.7$   & $38624$ & $4.4$   \\
\imdb   & $0.7$    & $5.9$  & \textbf{0.4}  & $26.6$  & $230.8$ & $230.7$ & --     & $35.3$   \\
\cancer & \textbf{5.5}    & $91.5$ & $9.5$  & $271.7$ & $38.8$  & $32.7$  & --        & $146.8$   \\
\bottomrule
\end{tabular}
}
\label{tab:efficiency}
 
\end{table}
%%%%%%%%%%%%%%%%%%%

\stitle{Scalability  \wrt graph size.}
Consider the overall performance of the   approaches from small-size, \ie~\aids to large-size \ie~\imdb, \cancer.
We find the following.
(1) Our \ourmtd  leverages the learned candidate matching nodes to directly prune unpromising search directions, which scales well to large-size graphs and also performs better on \ie~\imdb and \cancer from Table~\ref{tab:quality}.
(2) Combinatorial search-based algorithms can be extended to large-scale graphs with general performance due to aggressive relaxation (\hun and \vja ) or pruning strategies (\asbeam).
(3) Learning-based models add a bias to the predicted \ged values to reduce the discrepancy between the predicted and ground truth.
Their scalability  heavily relies on the ground truth produced by combinatorial search-based algorithms.
This is why they perform worse on \imdb and \cancer than \aids.
(4) The hybrid approach \gennas only completes \aids for less than $10$ nodes graphs and fails to scale to \imdb and \cancer.
This is because \gennas explores the entire space, \ie factorial scale and takes $\mathcal{O}(n^2d + d^2n)$ time for each search as explained in the introduction.

\stitle{Efficiency  \wrt  running time.}
The computational efficiency of eight approaches, evaluated over three real-life datasets, has been presented in Table~\ref{tab:efficiency}.
Due to the end-to-end learning, \simgnn and \genn achieve the best results and run in several microseconds to predict the \ged for one graph pair.
Though \ourmtd is slightly slower than the learning-based models, its running time is close to combinatorial search-based algorithms, and nearly  $10^4$ times faster than  the other hybrid approach \gennas.
This marked difference in performance underscores the significance of top-$k$ candidate node finds and the impact on the computational efficiency of \ourmtd.

%%%%%%%%%%%%%%%%%%% Ablation Studies
\begin{table}[tb!]
\centering
\renewcommand{\arraystretch}{0.95}
\caption{Ablation study. / \ourgnn  refers to replacing \ourgnn with GCN,  / LN refers to only removing  learning node matching,  / LG refers to only removing learning \ged, and / A* refers to only removing the matching refinement module. }
%\vspace{-2ex}
\setlength{\tabcolsep}{7pt}
\renewcommand{\arraystretch}{1.15}
\resizebox{0.95\columnwidth}{!}
{
\begin{tabular}{llccccc}
\toprule
 \multicolumn{1}{l}{Datasets}     & \multicolumn{1}{l}{Models}  & ACC $\uparrow$ & MAE $\downarrow$  & p@10 $\uparrow$   & $\rho$ $\uparrow$  \\ \midrule
\multirow{5}*{\aids}
& \ourmtd & \textbf{59.12} & \textbf{0.031}   & \textbf{0.486} & {0.844} \\
 &  / \ourgnn   & $54.38$ & $0.036$ & $0.485$ & $0.819$ \\
 &  / LN   & $32.13$ & $0.057$ & $0.397$ & $0.783$ \\
 &  / LG   & $58.09$ & $0.033$ & $0.469$ & $\textbf{0.858}$ \\
 &  / A*   & $9.28$ & $0.167$ & $0.244$ & $0.357$ \\

 \midrule
\multirow{5}*{\imdb}
 & \ourmtd & \textbf{44.80}   & \textbf{0.098}   & \textbf{0.509}   & $0.542$\\
 & / \ourgnn  & $41.43$   & $0.100$   & $0.493$   & $0.544$ \\
 & / LN   & $40.01$   & $0.102$   & $0.488$   & $0.541$ \\
 & / LG   & $42.99$   & \textbf{0.098}   & $0.505$   & \textbf{0.549} \\
 & / A*   & $30.16$   & $0.112$   & $0.428$   & $0.521$ \\

\midrule
\multirow{5}*{\cancer}
 & \ourmtd & \textbf{55.89}   & \textbf{0.040}   & \textbf{0.820}   & \textbf{0.729} \\
 & / \ourgnn &  $53.36$   & $0.042$   & $0.817$   & $0.699$ \\
 & / LN   & $45.85$   & $0.046$   & $0.790$   & $0.582$ \\
 & / LG   & $55.87$   & \textbf{0.040}   & $0.812$   & \textbf{0.729} \\
 & / A*   & $10.89$   & $0.104$   & $0.327$   & $0.146$ \\
\bottomrule
\end{tabular}
}  
\label{tab:ablation}  
\end{table}
%%%%%%%%%%%%%%%%%%%

%on the importance of each proposed component on the three real-life datasets, and the ablation  %\wrt different components.}

\stitle{Ablation study.}
We perform ablation studies to verify the effectiveness of each module and the results are reported in Table~\ref{tab:ablation}.
It is observed the performance drops dramatically if the matching refinement  module is removed. This is because multiple optimal node matchings typically exist from the node matching perspective of \ged. It is also noted that the performance considerably decreases if the structure-enhanced GNN (\ie~\ourgnn) is replaced by the vanilla one. It demonstrates that \ourgnn  successfully captures the combinatorial property of structure-dominant operations  and learns the powerful embeddings for approximate \ged computation.  Further, the two designed learning tasks (\ie LN, and LG)  are both helpful for
improving the solution quality of \ged, especially the learning node matching tasks (\ie LN). 
The ablation study is consistent with our analysis of the approximate \ged computation.

%%%%%%%%%%%%%%%%%%% topk selection
\begin{table}[tb!]
\centering
\renewcommand{\arraystretch}{0.95}
\caption{Performance evaluations \wrt different top-$k$. The metrics are calculated in the same way as those in Table \ref{tab:quality}. }
%\vspace{-2ex}
\renewcommand{\arraystretch}{1.15}
\resizebox{0.95\columnwidth}{!}
{
\begin{tabular}{lcccccc}
\toprule
 \multicolumn{1}{l}{Datasets}  & \multicolumn{1}{c}{ top-$k$} & ACC $\uparrow$ & MAE $\downarrow$ & p@10 $\uparrow$  & $\rho$ $\uparrow$ & Time $\downarrow$\\ \midrule
\multirow{6}*{\aids}
 & $5$  & $74.85$ & $0.018$ &  $0.586$ & $0.796$ &  \textbf{4.3} \\
 & $6$  & $84.10$ & $0.010$ &  $0.693$ & $0.859$ & $4.4$ \\
 & $7$  & $91.79$ & $0.005$ &  $0.787$ &  $0.910$ & $4.4$ \\
 & $8$  & $95.79$ & $0.002$ &  $0.846$ &  $0.940$ & $4.5$ \\
 & $9$  & $97.76$ & $0.001$ &  $0.912$ &  $0.959$ &  $4.8$ \\
 & 10 & \textbf{100.00} & \textbf{0.000}  & \textbf{1.000} &  \textbf{1.000}      & $5.2$  \\ \midrule
 
\multirow{6}*{\imdb}
 & $5$  & $38.57$ & $0.106$ &  $0.381$ &  $0.578$ &  \textbf{30.2} \\
 & $6$  & $39.40$ & $0.105$ &  $0.391$ &  $0.579$ & $35.3$ \\
 & $7$  & $40.97$ & $0.103$ &  $0.401$ & $0.582$ & $43.3$ \\
 & $8$  & $41.56$ & $0.102$ &  $0.400$ &  $0.582$ & $51.9$ \\
 & $9$  & $44.47$ & \textbf{0.100} &$0.412$ &  $0.587$ & $53.2$ \\
 & 10 & \textbf{45.05} & \textbf{0.100}  & \textbf{0.415} & \textbf{0.588}  & $55.3$ \\ \midrule

\multirow{6}*{\cancer}
 & $5$  & $5.01$  & $0.104$ &  $0.452$  & $0.678$  & \textbf{129.8} \\
 & $6$  & $7.37$  & $0.091$ &  $0.486$  & $0.709$   & $146.8$ \\
 & $7$  & $10.55$ & $0.079$ &  $0.543$  & $0.734$   & $153.1$ \\
 & $8$  & $14.37$ & $0.069$ &  $0.569$  & $0.758$  & $168.0$ \\
 & $9$  & $22.63$ & $0.058$ &  $0.615$  & $0.777$  & $176.7$ \\
 & 10 & \textbf{34.19} & \textbf{0.048} & \textbf{0.684}  & \textbf{0.794}   & $193.2$ \\

\bottomrule
\end{tabular}
} 
\label{tab:topk}
\end{table}
%%%%%%%%%%%%%%%%%%% in the same way as those in Table 2

\subsection{Performance \wrt top-$k$ selection.}
We also study the performance \wrt selecting different $k$ of \ourmtd on \aids, \imdb and \cancer datasets.
We conduct the following findings from Table \ref{tab:topk}. 

(1) Varying  $k$ in the experiments emphasize the trade-off between solution quality and time. That is setting larger $k$ could improve the approximate \ged solution quality, but the running time indeed increases mainly due to the large search space of \aslsa.
(2) \aids is a small dataset with the exact solutions, and \ourmtd also achieves the optimal solutions running in $5.2$ ms, when $k$ is set to $10$.
Note that, \ourmtd degenerates to \aslsa when all  nodes of $\mathcal{G}_2$ are selected as the candidate matching nodes.
(3) When $k$ is set to 6, 8 for datasets \imdb and \cancer, the evaluation metric  is worse than that  in Table \ref{tab:quality}.
In fact, $k$ is set to 10 in this test which achieves a smaller edit distance, and the ground-truth of \imdb and \cancer are further updated by these solutions.
Hence, the evaluation metrics are re-calculated, which produces worse metrics on \imdb and \cancer.
To encapsulate, the interplay of the value of $k$, offers a balance between computational scalability and accuracy of the approximate graph edit distance computation.

\subsection{ Analysis of assignment matrices of \ourmtd} 
\begin{figure}[tb!] 
\centering
\subfigure {
\label{fig:exp-sin-aan}
\includegraphics[width=.34\columnwidth]{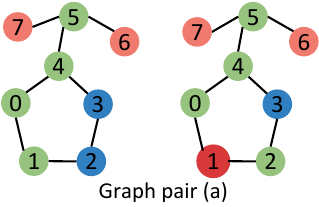}%
} 
\subfigure {
\label{fig:exp-sin-dblp}
\includegraphics[width=.22\columnwidth]{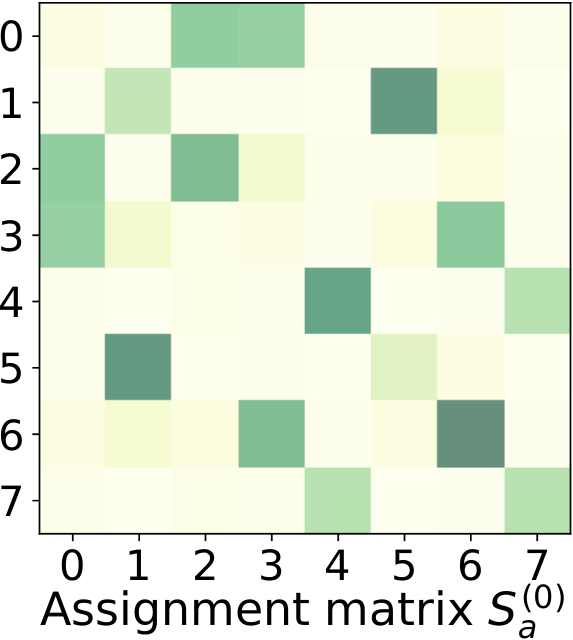}
} 
\subfigure{
\label{fig:exp-sin-acm}
\includegraphics[width=.22\columnwidth]{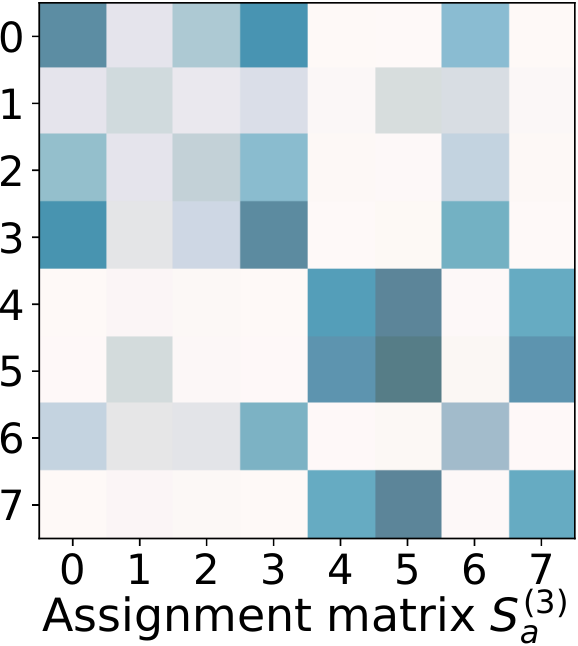}
}  

\subfigure {
\label{fig:exp-sin-mag}
\includegraphics[width=.34\columnwidth]{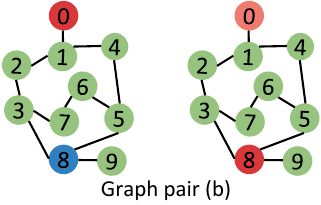}
}
\subfigure {
\label{fig:exp-sin-dblp}
\includegraphics[width=.22\columnwidth]{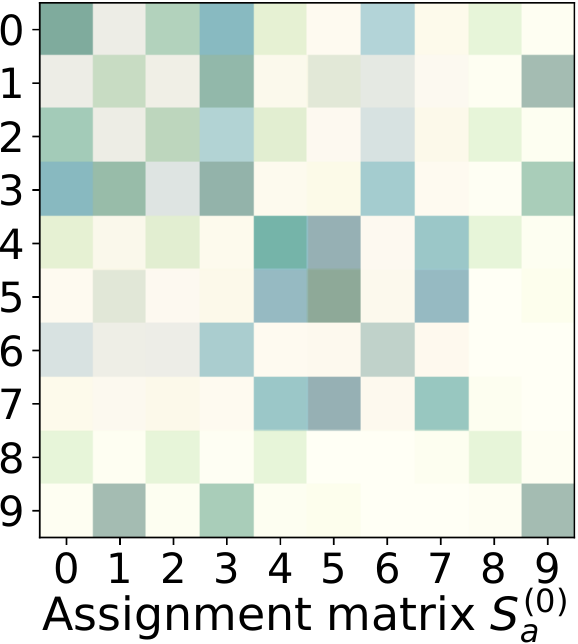}
} 
\subfigure {
\label{fig:exp-sin-acm}
\includegraphics[width=.22\columnwidth]{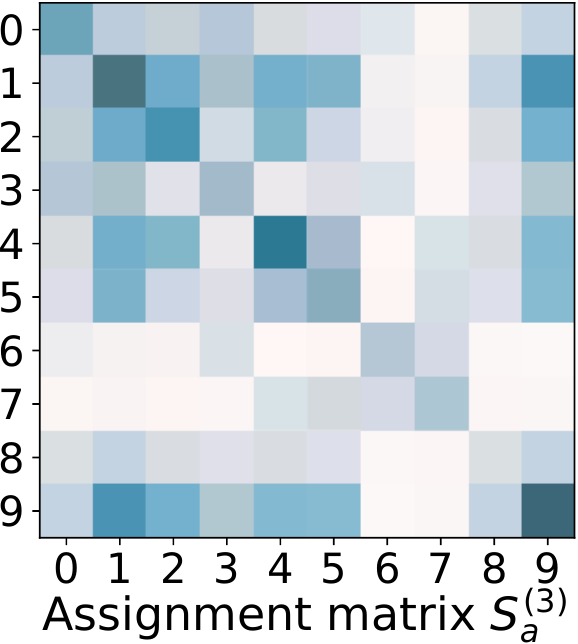}
}%\vspace{-2ex}
\caption{ Analysis of the  assignment matrices. The \ged of graph pairs (a) \& (b) are both equal to 2.}
\label{fig:exp-any}%\vspace{-2ex}
\end{figure}

we offer a visual representation of four  assignment matrices that encapsulate both local and high-order perspectives. 
These matrices, specifically pertaining to two pairs of graphs, have been portrayed as heatmap images, and are generated by \ourmtd on \aids in Fig.~\ref{fig:exp-any}.
From our observations, both the local and high-order views play a crucial role in the extraction of features tailored for node matchings. 
This is evident when considering specific node pairs, such as (6,6) and (7,7), from the graph pair labeled (a) (the top row of the assignment matrics $\mathbf{S_a}^{(0)}$ and $\mathbf{S_a}^{(l)}$).
We can see that the local and high-order views  both extract features for matching nodes, \eg the node pairs (6,6) and (7, 7) of the pair (a) (the top row).
Besides,  the assignment matrix $\mathbf{S_a}^{(l)}$  in the high-order view typically has a more powerful capacity to learn the node correspondence  compared to the local view $\mathbf{S_a}^{(0)}$. For example, $\mathbf{S_a}^{(0)}$  fails to capture the node pairs (1,1) of the pair (b), while $\mathbf{S_a}^{(l)}$ successfully learns it.
Thus, Fig.~\ref{fig:exp-any} shows the importance of using  local and high-order views to jointly learn the top-$k$ candidates rather than a single one.

%This is evident when considering specific node pairs, 
%the product of s operations on the Aids dataset.
%We demonstrate four assignment matrices from the local and high-order views, \wrt two graph pairs (plotted as heatmap images) generated by \ourmtd on \aids in Fig.~\ref{fig:exp-any}.

\section{Conclusion}
\label{sec-con}
We have presented a data-driven hybrid approach \ourmtd based on Graph Neural Networks (\ourgnn) and A* algorithms, which leverage the learned candidate matching nodes to prune unpromising search directions of \aslsa algorithm to approximate graph edit distance.
We have modeled it from a new perspective of node matching and combined the intrinsic relationship between \ged computation and node matching.
Besides, the design of our hybrid approach \ourmtd has been aware of the  two combinatorial properties involved in \ged computation: structure-dominant operations and multiple optimal node matching, to learn the matching nodes  from both local and high-order views. 
Benefiting from the candidate nodes, \ourmtd  has offered a balance between computational scalability and accuracy on the real-life datasets.
Finally, we have conducted extensive experiments on \aids, \imdb, and \cancer to demonstrate the effectiveness, scalability, and efficiency of combinatorial search-based, learning-based and hybrid approaches.

\begin{acks}
This work is supported in part by NSF of China under Grant 61925203  \& U22B2021. For any correspondence, please refer to Shuai Ma and Min Zhou. 
\end{acks}

\bibliographystyle{ACM-Reference-Format}
\balance
\bibliography{refs}

\end{document}